\documentclass[final]{cvpr}

\usepackage{times}
\usepackage{epsfig}
\usepackage{graphicx}
\usepackage{amsmath}
\usepackage{amssymb}
\usepackage[linesnumbered,boxed]{algorithm2e}
\usepackage{multirow}
\newcommand\blfootnote[1]{%
  \begingroup
  \renewcommand\thefootnote{}\footnote{#1}%
  \addtocounter{footnote}{-1}%
  \endgroup
}
\usepackage{subfig}

\usepackage[pagebackref=true,breaklinks=true,colorlinks,bookmarks=false]{hyperref}

\def\cvprPaperID{64} 
\def\confYear{CVPR 2021}

\begin{document}

\title{Multi-modal Bifurcated Network for Depth Guided Image Relighting}

\author{Hao-Hsiang Yang$^{1*}$, Wei-Ting Chen$^{1,2*}$, Hao-Lun Luo$^{3}$, and Sy-Yen Kuo$^{3}$\vspace{3mm}\\
$^{1}$ ASUS Intelligent Cloud Services, Asustek Computer Inc, Taipei, Taiwan\\
$^{2}$ Graduate Institute of Electronics Engineering, National Taiwan University, Taipei, Taiwan\\
$^{3}$ Department of Electrical Engineering, National Taiwan University, Taipei, Taiwan.\\{\tt\small (islike8399, jimmy3505090)@gmail.com, (r08921051, sykuo)@ntu.edu.tw}\\ \small\url{https://github.com/weitingchen83/NTIRE2021-Depth-Guided-Image-Relighting-MBNet}
}


\maketitle
\blfootnote{*Equally-contributed first authors.}

\begin{abstract}
Image relighting aims to recalibrate the illumination setting in an image. In this paper, we propose a deep learning-based method called multi-modal bifurcated network (MBNet) for depth guided image relighting. That is, given an image and the corresponding depth maps, a new image with the given illuminant angle and color temperature is generated by our network. This model extracts the image and the depth features by the bifurcated network in the encoder. To use the two features effectively, we adopt the dynamic dilated pyramid modules in the decoder. Moreover, to increase the variety of training data, we propose a novel data process pipeline to increase the number of the training data. Experiments conducted on the VIDIT dataset show that the proposed solution obtains the \textbf{1}$^{st}$ place in terms of SSIM and PMS in the NTIRE 2021 Depth Guide One-to-one Relighting Challenge.
\end{abstract}

\section{Introduction}

\begin{figure}[t!]
	\centering
	\subfloat[]{\includegraphics[width=0.22\textwidth]{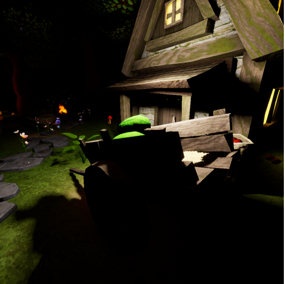}}
    \hspace{0.01em}
	\subfloat[]{\includegraphics[width=0.22\textwidth]{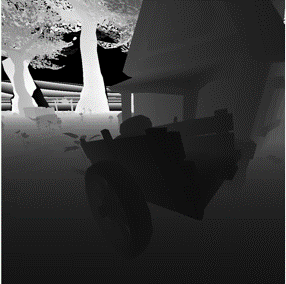}}
    \hspace{0.01em}
	\subfloat[]{\includegraphics[width=0.22\textwidth]{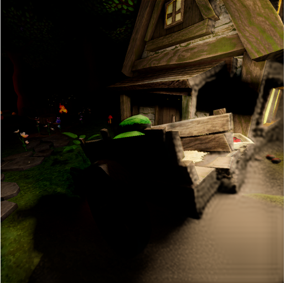}} 
	 \hspace{0.01em}
	\subfloat[]{\includegraphics[width=0.22\textwidth]{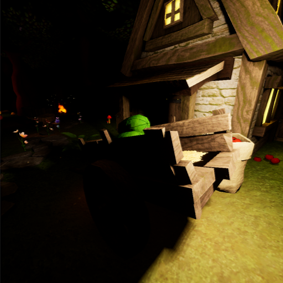}}
	\caption{Example of depth guided image relighting. (a): Original input image. (b): Corresponding depth map. (c): Relighted image by our method. (d): Ground truth.}
\label{fig:fig1}
\end{figure}

Given an image, image relighting aims to relight this image into another image with different ambient conditions. In the NTIRE 2021 Depth Guide One-to-one Relighting Challenge, the depth maps are provided. We plot examples including the original image, the corresponding depth map, the relighted image by the proposed method, and the ground truth in \figref{fig:fig1}. As shown in \figref{fig:fig1}, there are two inherent challenges for image relighting. First, to generate the image with different ambient conditions, it is necessary to generate shadows into the relighted image. Second, similarly, the shadow from the original image needs to be removed. For example, in \figref{fig:fig1} (c), although the region shadows are removed, the texture of the grass cannot be recovered appropriately. 

Previously, many methods \cite{debevec2000acquiring,matusik2004progressively,reddy2012frequency} based on developing visual priors or capture properties of relighted images have achieved impressive performance. Recently, some deep learning-based methods \cite{xu2018deep,puthussery2020wdrn,elhelou2020aim,yang2021S3Net} are proposed without explicit inverse rendering steps for estimating scene properties. However, these methods do not consider complex scenes and various ambient conditions. Moreover, in NTIRE 2021 Depth Guide One-to-one Relighting Challenge, there are extra difficulties needed to be addressed. First, the number of training data given in the VIDIT \cite{helou2020vidit} dataset is 300, which is not enough to train the network. To address these issues, our network is based on the backbone pre-trained from ImageNet. Furthermore, we leverage extra data from Track II: Any-to-Any relighting \cite{yang2021S3Net} and develop a strategy that can generate new image pairs to increase the number of the training data and the robustness of the network. 

Second, depth and image features should be effectively extracted and fused. Depth maps and images contain the different attributes of features. Depth maps present spatial information while the images provide texture, light cues, and dark cues. Thus, we refer to the methods proposed in the RGB-D salient object detection \cite{pang2020hierarchical,chen2020progressively,zhang2020asymmetric} and propose a multi-modal bifurcated network to deal with this issue. This network contains encoder and decoder parts. Our encoder applies two branches without sharing weight to extract features, respectively. Additionally, we apply the dynamic dilated pyramid module to effectively integrate two features. In the decoder parts, we gradually magnify the feature maps and recover the image. Motivated by the U-Net \cite{ronneberger2015u,yang2019wavelet}, we apply skip connection to connect the feature maps with identical size from the encoder and decoder parts to obtain better relighted images.

We make the following contributions in this paper:
\begin{enumerate}
\item The multi-modal bifurcated network is proposed for depth guided image relighting. This structure can extract image and depth features by two branches. Then, these two features are fused by dynamic dilated pyramid modules effectively.  
\item We propose a new strategy to leverage additional images from Track II and construct more input-output pairs as the training data.
\item Several experiments performed on the VIDIT \cite{helou2020vidit} dataset demonstrate that our solution achieves 
the \textbf{1}$^{st}$ place in terms of MPS and SSIM in the NTIRE 2021 Depth Guided One-to-one Relighting Challenge.
\end{enumerate}


\section{Related Works}
Two tasks are very similar to depth guided image relighting: RGB-D saliency object detection and image relighting. In this section, we briefly describe several works related to these tasks.
\subsection{RGB-D Salient Object Detection}
Salient Object Detection (SOD) aims to imitate the human
visual system and detect certain regions or objects that attract human attention. RGB-D SOD is known as combining the extra depth maps to fulfill salient detection. Some models \cite{chen2020progressively,zhang2020asymmetric} extract features from images and depth maps independently, and conducts feature maps fusion of the two modalities in the decoder. In \cite{zhang2020asymmetric}, Asymmetric Two-Stream Architecture (ATST) is proposed which considers the inherent differences between the RGB and the depth data for the salient detection. In \cite{chen2020progressively}, Guided Residual (GR) blocks are proposed to feed the RGB image and the depth image alternately to reduce the mutual degradation. They also address progressive guidance in the stacked GR blocks within each side-output to remedy the false detection and the missing parts. On the other hand, in \cite{zhao2020single}, the single stream network is designed that concatenates the RGB image and depth image across the channel dimension and directly applies the depth map to guide both the early fusion and the middle fusion between the RGB information and the depth information, which saves the feature encoder of the depth stream. Additionally, in \cite{chen2021rgb}, authors propose the RD3D that 3D convolutional neural networks are introduced to address the RGB-D SOD. This network adopts the progressive fusion involving both the encoder and the decoder stages.
Since the multi-modal fusion methods from RGB-D SOD can be leveraged for the depth-guided image relighting, in this paper, the proposed MBNet is based on the HDFNet \cite{pang2020hierarchical}.
\begin{figure*}[t!]
  \centering
\includegraphics[width=0.8\textwidth]{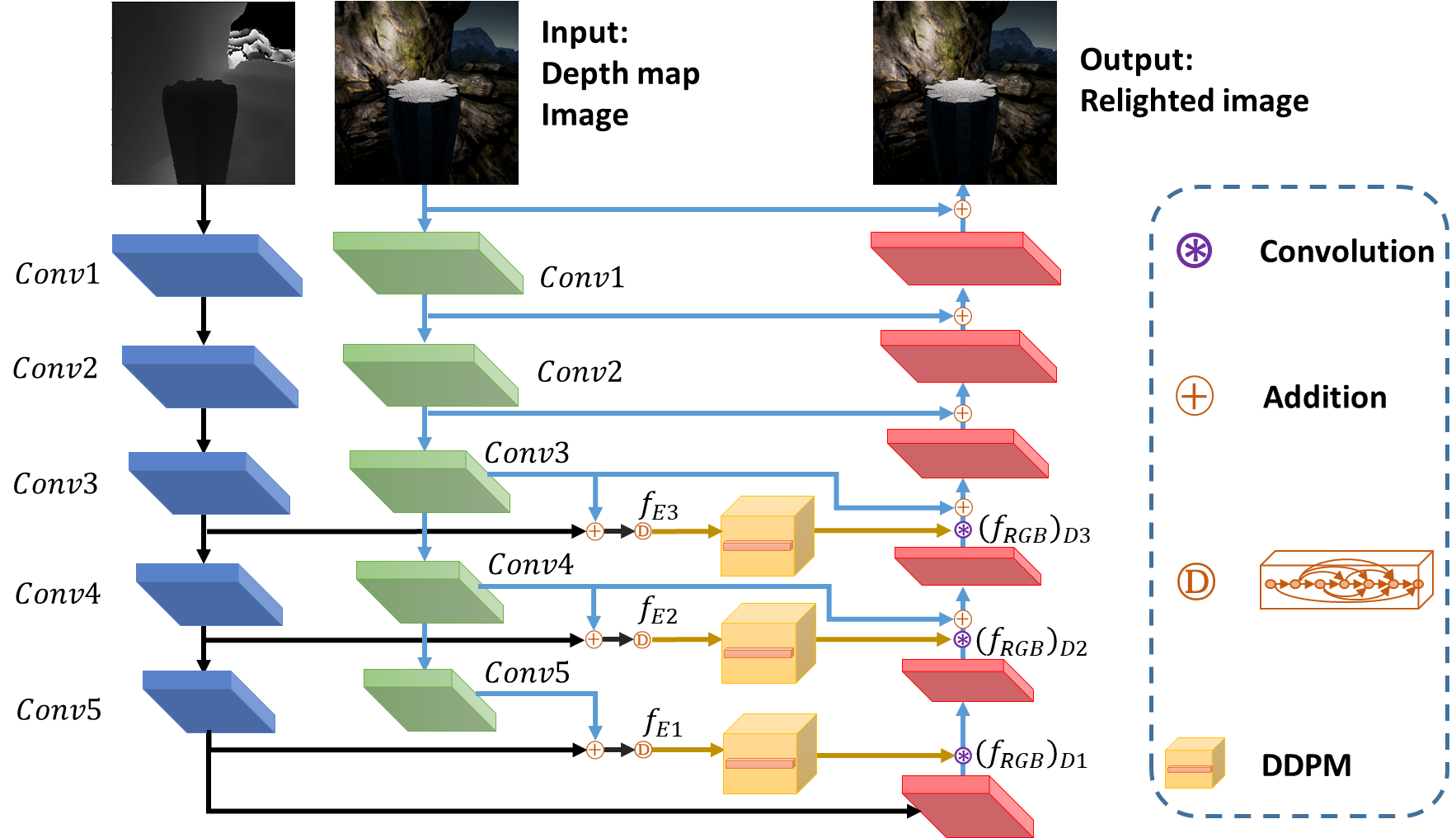}

    \caption{{The architecture of the proposed multi-modal bifurcated network. The network consists of two streams as encoder parts: depth stream and RGB-image stream. We use the dense architecture ,DDPM and skip connection for better feature extraction as decoder parts.
}}
\label{fig:architecture}
\end{figure*}

\subsection{Image Relighting}
The algorithms for image relighting can be categorized as the physical-based method and the deep learning-based method. Traditional methods \cite{debevec2000acquiring,matusik2004progressively,reddy2012frequency} depend on the physical observation to further estimate the ambient conditions, reflectance, and lighting of the scene in the image and then re-render this scene by another illumination setting.
In \cite{matusik2004progressively}, an algorithm treating the complex scene as a linear system that transforms the original light into the reflected light is proposed. This algorithm progressively refines the approximation of the reflectance field until the required precision is reached. In \cite{reddy2012frequency}, authors prove that the light transport of diffuse scenes under the spatially varying illumination can be decomposed into the direct, near-range and far-range transports. They separate these three components in the frequency space.
On the other hand, many deep learning-based methods for image relighting \cite{xu2018deep,puthussery2020wdrn,yang2021S3Net} are proposed. Image relighting can be seen as an image-to-image translation problem. Several low-level image processing problems like haze/smoke removal \cite{yang2020net,chen2018color,chen2020pmhld}, underwater enhancement \cite{yang2021LAFFNet}, reflection removal \cite{tsai2018efficient}, 
image deraining \cite{yang2020wavelet} and image desnowing \cite{chen2020jstasr} are very similar to image relighting. Generally speaking, the encoder-decoder structure like U-Net \cite{ronneberger2015u,yang2019wavelet} can deal with these tasks. Some methods that deal with the ambient conditions especially for the image relighting are developed. Gray loss described in \cite{puthussery2020wdrn} can drive the network to learn the illumination gradient in target domain images. Xu \textit{et.al} propose a CNN-based method \cite{xu2018deep} to relight a scene under a new illumination based on five images captured under a pre-defined illumination setting. This method tries to estimate a non-linear function that generalizes the estimation from the optimal sparse samples.

\section{Proposed methods}
\subsection{Overall Neural Network}
As shown in \figref{fig:architecture}, to leverage the depth information for depth guided image relighting task effectively, we refer the dual-stream bifurcated architecture in \cite{pang2020hierarchical} as the backbone. Our network consists of two streams to extract the depth and the image features. We apply the two ResNet 50 \cite{he2016deep} pre-trained from the ImageNet as the backbones. In the network design, the depth and image features from three intermediate layers are fused to achieve the representative features. We fuse the features from the conv3, the conv4 and the conv5 layers to balance the effectiveness and the efficiency of the network. Specifically, for the features in the shallower layers, they are generally noisier and the high-resolution of these features may increase the computational burden. However, the features in the conv3 to the conv5 may still preserve the valid information \cite{pang2020hierarchical} and with lower resolution. To fuse these two features with multiple receptive fields, we leverage the densely connected architecture to generate the combined features with the fruitful texture and the structural information. Then, these features are fed to the Dynamic Dilated Pyramid Module (DDPM)\unskip~\cite{pang2020hierarchical} that can generate a more discriminative result. We will describe this module in the following section. The output of the DDPM is combined with the output of the decoder by convolving with the multi-scale convolution kernels\unskip~\cite{chen2019pms,chen2020jstasr,ren2016single}. In the decoder part, similar to the U-net \cite{yang2019wavelet,ronneberger2015u}, we gradually magnify the feature maps and implement the skip connection to concatenate the identical size feature maps. Furthermore, we make our network learn the residual \cite{he2016deep} instead of the whole images. That is, the final output is the difference between the original image and the relighted image.

\subsection{Dynamic Dilated Pyramid Module}

\begin{figure*}[t!]
  \centering
\includegraphics[width=0.8\textwidth]{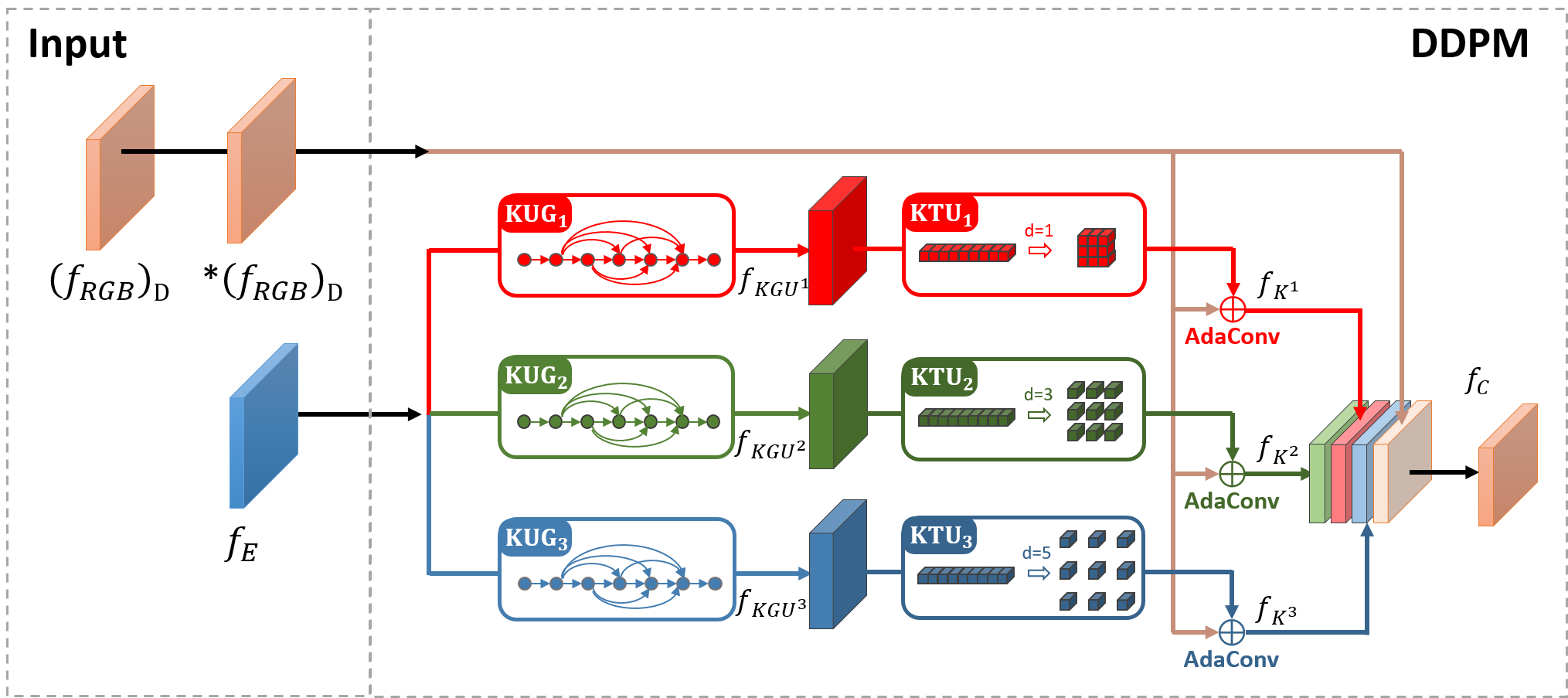}
    \caption{{The architecture of the dynamic dilated pyramid module (DDPM). There are two modules in the network, that is, the kernel generation units (KGUs) and the kernel transformation units (KTUs).}}
\label{fig:architecture_DDPM}
\end{figure*}

In this part, we illustrate the detail of the DDPM. As shown in \figref{fig:architecture_DDPM}, the input of DDPM is the fused feature $f_{E}$ and the feature $(f_{RGB})_{D}$ from encoders, respectively. First, the $(f_{RGB})_{D}$ passes through the convolution kernel to reduce the dimension of the feature which is termed as $^{*}(f_{RGB})_{D}$. Second, the kernel generation units (KGUs) \cite{pang2020hierarchical} are adopted on the fused feature $f_{E}$ to generate different weight tensors (i.e, $f_{KGU}^{1}$, $f_{KGU}^{2}$, $f_{KGU}^{3}$) which can cover three square neighborhoods (i.e., $3\times3$, $7\times7$, and $11\times11$) . It is noted that KGUs are with four densely connected layers \cite{huang2017densely} which can improve the feature propagation and feature reuse effectively. Then, we leverage the kernel transformation units (KTUs) to yield regular convolution kernels with various dilation rate (i.e., 1, 3, 5) by reorganizing kernel tensors and inserting various numbers of zeros. Then, the three parallel output features combine with the $^{*}(f_{RGB})_{D}$ by the convolution kernels, respectively. We term these combined features as $f_{K}^{1},f_{K}^{2},f_{K}^{3}$. Finally, we combine $f_{K}^{1},f_{K}^{2},f_{K}^{3}$ and $^{*}(f_{RGB})_{D}$ to generate the output of the DDPM $f_{C}$.

\subsection{Extra Data Usage}
\begin{figure*}[t!]
	\centering
	\subfloat[]{\includegraphics[width=0.24\textwidth]{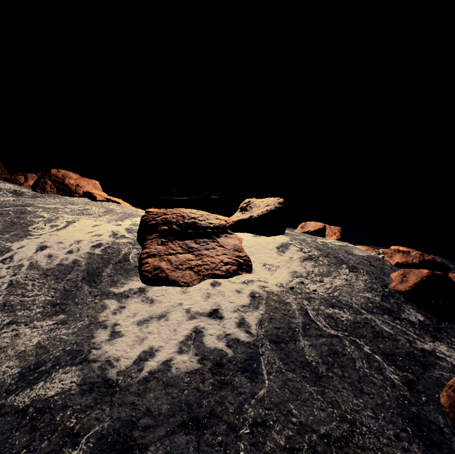}}
    \hspace{0.01em}
	\subfloat[]{\includegraphics[width=0.24\textwidth]{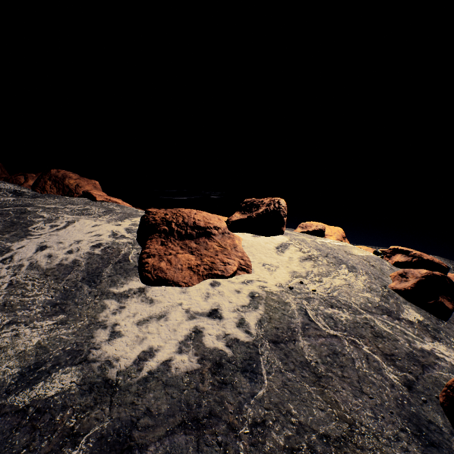}}
    \hspace{0.01em}
	\subfloat[]{\includegraphics[width=0.24\textwidth]{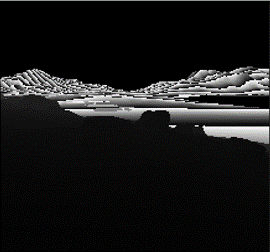}}
	 \hspace{0.01em}
	\subfloat[]{\includegraphics[width=0.24\textwidth]{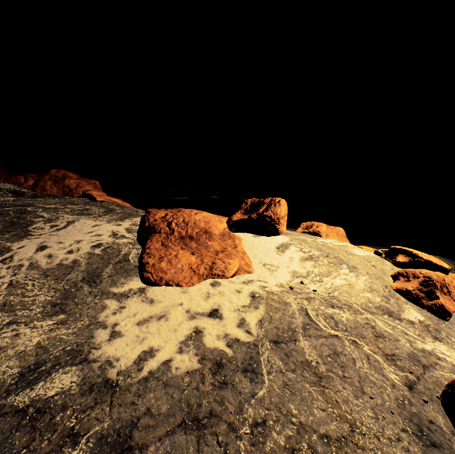}}
	\caption{Additional images are used in the training phase. (a) Original image (6500-N). (b) Different illuminating angle image (6500-NE). (c) Guided depth map. (d) Output image.}
	\label{fig:aug}
\end{figure*}

In this section, we propose two strategies to increase the training data so that our model can learn the mapping function of the depth guided image relighting robustly. In this challenge, the input image (6500-N) is with 6500K color temperature and the north illumination angle while the output image (4500-E) is with 4500K color temperature and with the east illumination angle. In order to improve the robustness of our network, we leverage the images in Depth Guided Image Relighting: Track II Any-to-Any relighting \cite{yang2021S3Net}. This track provides the images with various illumination temperature and different illumination angles. Specifically, we apply the images with the different illumination angle (6500-NE) but the identical color temperature as the input and the ground truths are the same ones (4500-E). As shown in \figref{fig:aug}, the additional image (6500-NE) is very similar to the original one (6500-N). With this additional data, the model can understand the direction information comprehensively.

\begin{figure*}[t!]
	\centering
	\subfloat[]{\includegraphics[width=0.16\textwidth]{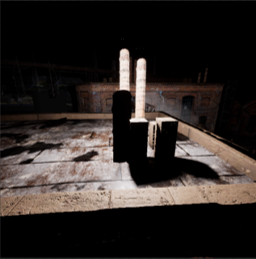}}
    \hspace{0.01em}
	\subfloat[]{\includegraphics[width=0.16\textwidth]{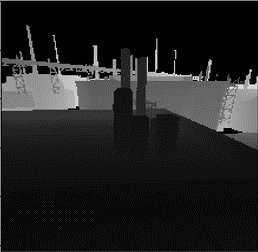}}
    \hspace{0.01em}
	\subfloat[]{\includegraphics[width=0.16\textwidth]{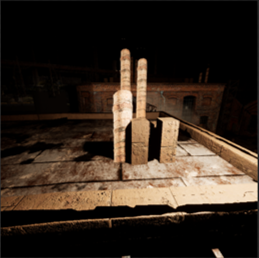}}
	 \hspace{0.01em}
	\subfloat[]{\includegraphics[width=0.16\textwidth]{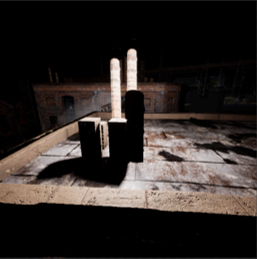}}
	\hspace{0.01em}
	\subfloat[]{\includegraphics[width=0.16\textwidth]{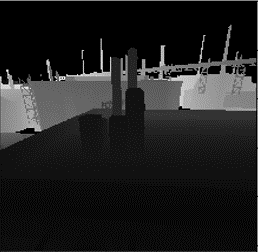}}
	\hspace{0.01em}
	\subfloat[]{\includegraphics[width=0.16\textwidth]{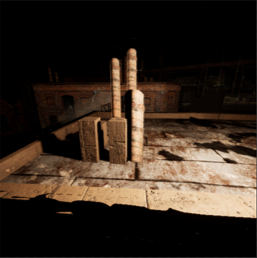}}
	\caption{Using (4500-W) images to increase the diversity of the training data. (a): Original input image. (b): Original depth map. (c): Output image. (d): Flipped image of (a). (e): Flipped depth map of (b). (f): Flipped image of (4500-W).}
\label{fig:aug2}
\end{figure*}

Moreover, We adopt additional images which contain the same scene but with the different illuminating angle (4500-W). We can flip horizontally the image with the west illumination angle to achieve the new image with the east illumination angle. Thus, we develop a new strategy to further increase training data and illustrate it in \figref{fig:aug2}. As shown in \figref{fig:aug2} (d) and \figref{fig:aug2} (e), we flip the original RGB (6500-N) images and the corresponding depth map horizontally. The horizontally flipped image of (4500-W) is the output of the flipped input (6500-N). With this operation, the training data can be increased.

\subsection{Loss Functions}
In this paper, we leverage three loss functions to measure the differences between the relighted images and the ground truth. The three loss functions are the Charbonnier loss \cite{barron2019general}, the SSIM loss \cite{zhao2016loss}, and the perceptual loss \cite{johnson2016perceptual}. The Charbonnier loss can be expressed as:
\begin{equation}
L_{Cha}(x, \hat x) = \frac{1}{T}\sum_{i}^{T}\sqrt[]{(x_i-\hat{x_i})^2+\epsilon ^2}
\label{eq:res}
\end{equation}
where $x$ and $\hat x$ are the ground truth and relighted images, respectively. $e$ is a tiny constant for the stable and the robust convergence.

The second loss function is the SSIM loss function. SSIM loss is expressed as:
\begin{equation}
  {L_{SSIM}}(x,\hat x) = -SSIM(x,\hat x)
  \label{eq:SSMI}
\end{equation}
SSIM loss is beneficial to reconstruct local structures and details.

Finally, the perceptual loss is written as:

\begin{equation}
L_{Per}(x,\hat{x})= |(VGG(x)-VGG(\hat{x})|
\label{eq:res}
\end{equation}
where $VGG$ means the VGG19 network \cite{simonyan2014very}.  In our work, we use the features from conv3-3 layer.
The overall loss function containing three terms is expressed as:
\begin{equation}
L_{Total}= \lambda_{1} L_{cha}+ \lambda_{2} L_{SSIM}+ \lambda_{3} L_{Per}
\label{loss}
\end{equation}
where $\lambda_{1}$, $\lambda_{2}$ and $\lambda_{3}$ are weights to control the final objective functions. These three weights are empirically adjusted as hyper-parameters. 

\section{Experiments}

\subsection{Experimental Setting}
In the NTIRE 2021 Depth Guide One-to-one Relighting Challenge, the novel Virtual Image Dataset for Illumination Transfer (VIDIT) \cite{helou2020vidit} is provided as the training and the validation data. This dataset consists of 390 various scenes that are captured at 40 different illumination conditions including 8 different azimuthal angles and five color temperatures such as 2500K, 4500K, etc. Furthermore, the corresponding depth maps are provided. In track I - depth guided one-to-one relighting, the input images are depth map and a pre-defined illumination condition $\theta$ = North, temperature = 6500K (e.g., (6500-N)) and the output image is set at a different illumination setting $\theta$ = East, temperature = 4500K. Though in track I, only two conditions of images are used as the input and the output pairs, it is allowed to utilize the extra data to improve the accuracy of the model. During the training and the evaluation phases, the image size is 1024 $\times$ 1024, and we do not use any data augmentation like random flip and random crop. The Adam optimizer \cite{kingma2014adam} is utilized with a batch size of 3 to train the network. We train the network for 200 epochs with the momentum $\beta_{1}$ = 0.5, $\beta_{2}$ = 0.999. The learning rate is initialed as $10^{-4}$ and divided by ten after 50 epochs. The $\lambda_1$, $\lambda_2$ and $\lambda_3$ in \eqref{loss} are set as 1, 1.1 and 0.1, respectively. We perform our experiments on a single Nvidia V100 graphic card and the PyTorch platform. We spend about 11 hours finishing the model training. In the testing phase, we take 2.8867 seconds to predict a single image. The source code will be available in our project page.

\begin{table}[ht]
\small
    \centering
    \caption{{The ablation experiment of applying the different data and the residual learning.}}
\begin{tabular}{l|cc}
\hline
Description   & PSNR & SSIM \\ \hline\hline
Baseline &     18.0215  &  0.6834 \\
+ Extra data  &  18.9677    &  0.7103    \\
+ Extra data + Residual learning  &   \textcolor[rgb]{ 1,  0,  0}{19.3558}   &  \textcolor[rgb]{ 1,  0,  0}{0.7175}
\end{tabular}
\label{tab:ablation}
\end{table}

\begin{table}[ht]
\small
  \centering
  \caption{Relighted results by some state-of-the-art RGB-D salient object methods.}
    \begin{tabular}{l|cc}
    \hline
    Methods & PSNR & SSIM \\ \hline \hline
    ATST \cite{zhang2020asymmetric}  & 18.2678 & 0.665 \\
    DANET \cite{zhao2020single} & 18.3341 & 0.6805 \\
    RD3D \cite{chen2021rgb} & 17.7763 & 0.6668 \\
    PGAR \cite{chen2020progressively}  & 17.6436 & 0.6748 \\
    Ours  & \textcolor[rgb]{ 1,  0,  0}{19.3558} & \textcolor[rgb]{ 1,  0,  0}{0.7175} \\
    \end{tabular}%
  \label{tab:addlabel}%
\end{table}%

\begin{figure*}[t]
  \centering
\includegraphics[width=\linewidth]{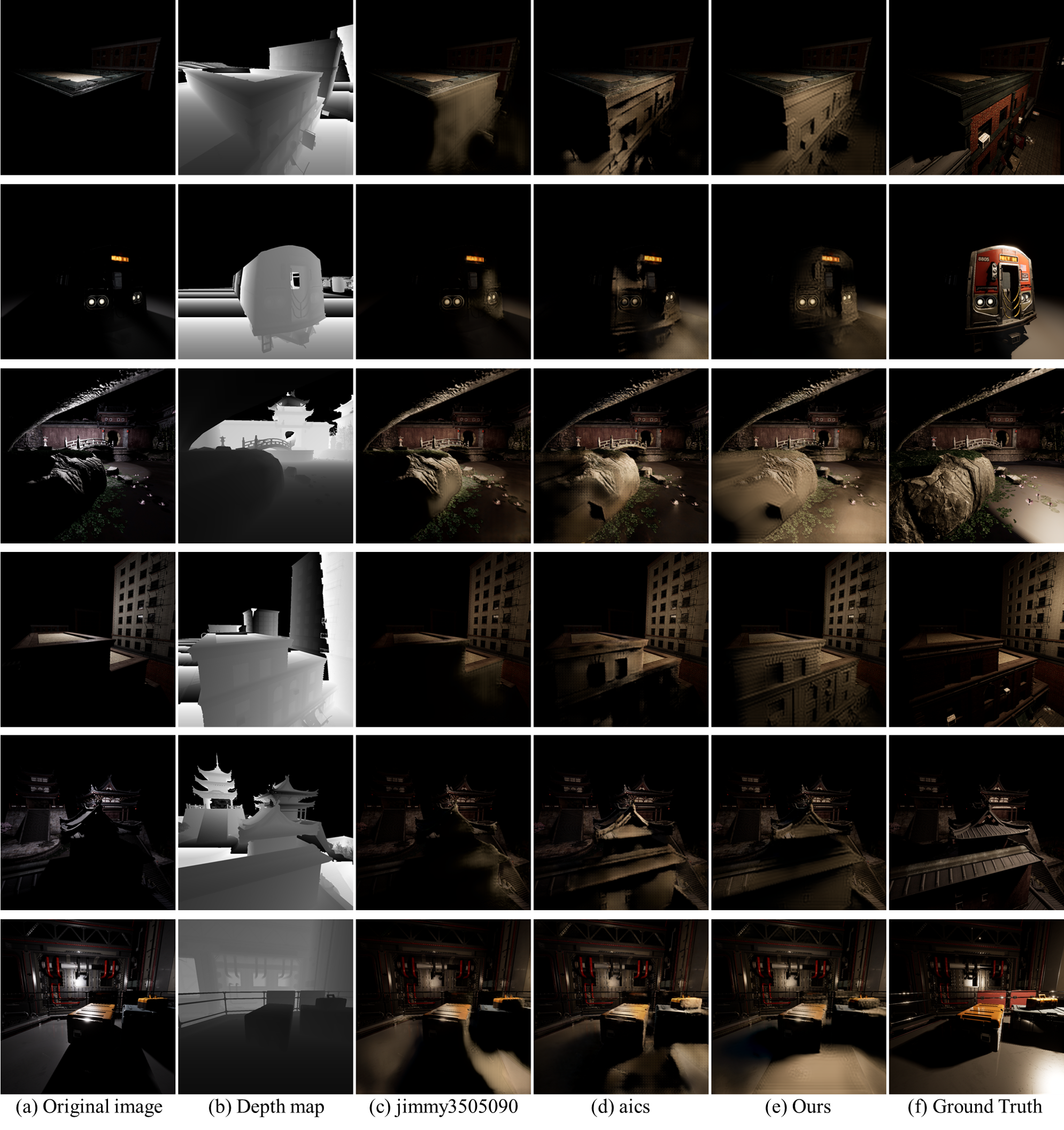}
    \caption{{Visual comparison for the relighted results recovered by the MBNet and other solutions.
}}

\label{fig:vis}
\end{figure*}

\begin{table*}[tbp]
  \centering
  \caption{The average SSIM, PSNR, MPS and LPIPS  of some submissions over NTIRE 2021 depth guided image relighting validation and testing dataset.}
    \begin{tabular}{r|cc|cccc}
    \multicolumn{1}{l|}{User name} & \multicolumn{2}{c|}{Validation} & \multicolumn{4}{c}{Testing} \\
    \hline
          & SSIM  & PSNR  & MPS   & SSIM  & LPIPS & PSNR \\ \hline
    auy200& 0.6937&18.4492&0.7620 & 0.6874 & 0.1634 & 18.8358 \\
    aics &0.7069&19.1026& 0.7601 & 0.6799 & 0.1597 & 18.8639  \\
    lifu &0.7104&19.0048& 0.7600 & 0.6903 & 0.1702 & 19.8645 \\
    jimmy3505090 &0.7069&18.2937& 0.7551 & 0.6772 & 0.1670 & 18.2766\\
    DeepBlueAI & 0.6757&18.6800&0.7494 & 0.6879 & 0.1891 & 19.8784\\
    Ours &0.7175 & 19.3558& 0.7663 & 0.6931 & 0.1605 & 19.1469 \\          
    \end{tabular}%
  \label{tab:addlabel}%
\end{table*}%

\subsection{Ablation Experiments}
We conduct the ablation experiment to verify that each module applied in this paper can benefit the proposed relighting network. In all experiments, the image size is set as 1024$\times$ 1024. We test each module and report the results under the validation set. We select the peak signal-to-noise ratio (PSNR) and the structural similarity (SSIM) as objective metrics for the quantitative evaluation. Overall, the ablation studies consist of three different experimental scenarios: 1) We apply the original training data to train the MBNet as the baseline. 2) We apply the method described in Section 3 to increase the training data to train our MBNet. 3) We increase the training data and also apply the residual learning \cite{he2016deep} strategy. We summary the ablation experiments in Table \ref{tab:ablation}. One can see that both PSNR and SSIM scores of setting 2 are increased compared with setting 1. It can show that increasing training data is beneficial for better performance and robustness. Additionally, compared with setting 2, the performance of setting 3 is improved effectively. It demonstrates that the residual learning can further improve the accuracy of the relighting. 

\subsection{Comparison with State-of-the-art Methods}
First, we compare the MBNet with four state-of-the-art RGB-D SOD methods including ATST \cite{zhang2020asymmetric}, DANET \cite{zhao2020single}, RD3D \cite{chen2021rgb} and PGAR \cite{chen2020progressively} as described in the Section 2. Note that, we use the same training set to train these methods. We replace the final convolutional layers of RGB-D SOD networks, so their output is three-channel tensors as relighted images. As shown in Table \ref{tab:addlabel}, the MBNet outperforms other methods with a large margin. Our method achieves the best performance on both PSNR and SSIM, which surpasses the second place 1.02 dB and 0.037 in SSIM.

Furthermore, we report the performances of some submissions in the NTIRE 2021 Depth Guide One-to-one Relighting Challenge \cite{elhelou2021ntire}. The performance is evaluated on the validation and the test dataset and the results are shown in Table \ref{tab:addlabel}. Additionally, The Mean Perceptual Score (MPS) is used for final evaulation. The MPS is defined as the average of the normalized SSIM and LPIPS \cite{zhang2018perceptual} scores. The MBNet produces moderate quality outputs and the \textbf{$1^{st}$} place performance in both MPS and SSIM metrics. We also plot some relighted images generated by our method and other participants in \figref{fig:vis}. Compared to other methods, our images can remove more shadows and present clear outlines of objects, though some results are not satisfactory.

\section{Conclusion}
In this paper, to address depth guided image relighting, we develop the multi-modal bifurcated network. This network extracts both depth and image features by a dual bifurcated backbone. To fuse multi-modal features, the dynamic dilated pyramid module is introduced. This module contains densely connected layers and multi-scale kernels to fuse and refine features from a dual bifurcated backbone. Furthermore, to improve the robustness and performance, we propose a new strategy to increase the training image pairs by leveraging extra images. Several experiments implemented on the novel VIDIT \cite{helou2020vidit} dataset proves that our solution achieves the \textbf{$1^{st}$} place in terms of MPS and SSIM in the NTIRE 2021 Depth Guided One-to-one Relighting Challenge.

\section{Acknowledgement}
We thank to National Center for High-performance Computing (NCHC) for providing computational and storage resources.

{\small
\bibliographystyle{IEEEtran}
\bibliography{egbib}
}

\end{document}